# Data Strategies for Fleetwide Predictive Maintenance

**David Noever**
david.noever@peoplete.com
Senior Technical Fellow
PeopleTec, Inc.
Huntsville, AL, USA

## ABSTRACT

For predictive maintenance, we examine one of the largest public datasets for machine failures derived along with their corresponding precursors as error rates, historical part replacements and sensor inputs. To simplify the time-accuracy comparison between 27 different algorithms, we treat the imbalance between normal and failing states with nominal under-sampling. We identify 3 promising regression and discriminant algorithms with both higher accuracy (96%) and twenty-fold faster execution times than previous work. Because predictive maintenance success hinges on input features prior to prediction, we provide a methodology to rank-order feature importance and show that for this dataset, error counts prove more predictive than scheduled maintenance might imply solely based on more traditional factors such as machine age or last replacement times.

## INTRODUCTION

Successful predictive maintenance is challenging not only because failures can prove multi-factorial but also because maintenance forecasters often lack good training data. As noted from a gap analysis in Ref. 1, the wider sharing of maintenance data offers the clearest path forward to discover better machine learning methods and improved maintenance decision-making. For aircraft applications, the main public reference datasets are the University of Cincinnati bearing faults in Ref. 2 and NASA's turbofan engine failures in Ref. 3. Neither of these deal with complex schedules for replacement parts or non-vibrational precursors as fleet managers might encounter. In a previous paper, we examined the large-scale deployment of algorithm families to solve helicopter sensor problems and detect a single bearing fault using a machine learning suite in Ref 4. Here we extend this approach to the complex and more general case of predicting fleetwide failures using multiple input sensors, time lags, maintenance logs, and component error rates.

## METHODOLOGY

We employ a novel synthetic dataset as one of the newest and largest public repositories for predictive maintenance in Ref. 5. While idealized their problem set offers prototypical issues faced in maintaining helicopter fleets, particularly where differently aged transmission or engine parts may provide complex spectra for vibration, oil pressure, temperatures (voltages) and rotation speeds. The dataset includes 100 machines of differing ages, each with four major components and five error types. Each machine has a distinct maintenance history with both unscheduled and scheduled replacements, subcomponent errors and finally overall machine failures. Given a generic sensor suite that measures vibration, rotation, voltage and pressure over time, the problem statement calls for fleetwide failure predictions. Although the data originate synthetically, it also offers an advantage as a defined, clean set for public (and competitive) algorithmic study. The data also offer significant depth and breadth. Its scope includes 3.2 million sensors samples (876,101 time-stamps for 4 sensor readings) for 100 machines with different ages, replacement histories and remaining useful lives. As expected for such failures within a fleet, the distribution of nominal events overwhelms both errors and maintenance events. The dataset suggests such outliers occur fewer than one in a thousand. Similarly, machine or component failures (the prediction target) occur even less (with frequency approximately 1/4000). The dominance of nominal behavior thus requires careful consideration either to re-balance the data or otherwise to devise useful metrics to score a given algorithm's performance. In the absence of these steps, any convergence on a zero-failure prediction would show nearly 100 percent accuracy and yet remain useless as a model for actual failures.

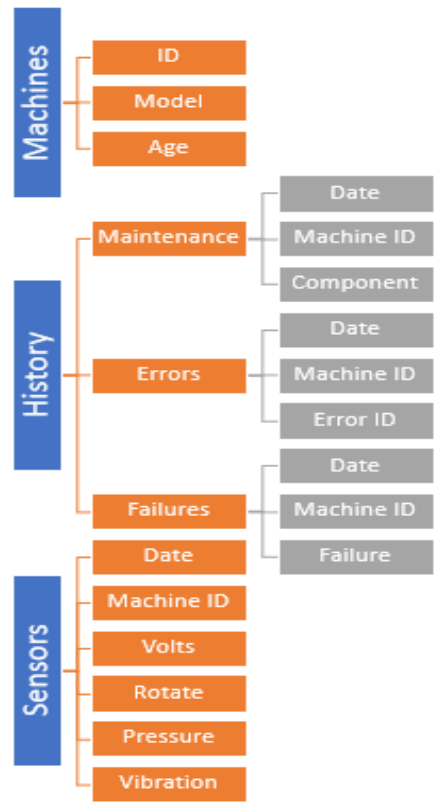

**Figure 1 Data Structures Needed for Predictive Maintenance**

## DATASET PREPROCESSING AND FEATURE EXTRACTION

Regardless of whether choosing just one or multiple algorithm(s), most benchmarks have improved performance when selecting better features. Feature selection here refers to the unique combination of historical states that accompany a given machine's failure. In Ref. 5 (and typical for many data science problems), the bulk of the effort focused on the need to transform and merge the five diverse data streams

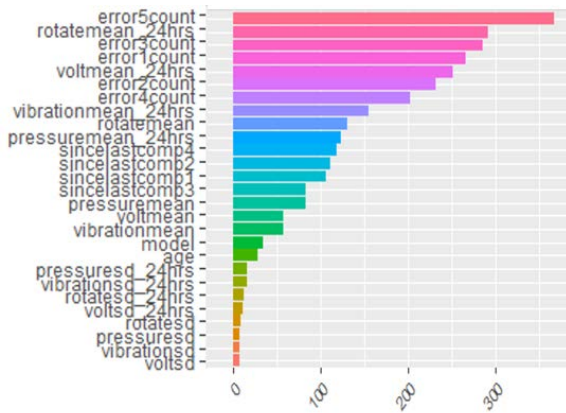

Figure 3 Rank Order of Input Variable Importance for Predicting Future Maintenance

(errors, sensors, maintenance, failures and machine metadata or identifiers). Feature selection included generating 27 factors to highlight outliers using the traditional mean and standard deviations for each sensor and two different time lags (fast=3 hours and

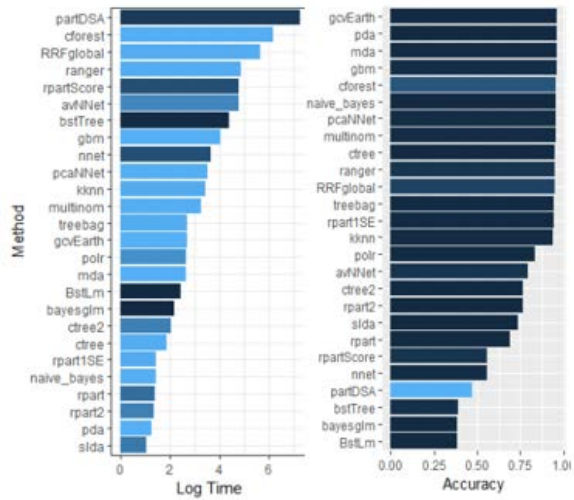

Figure 2 Algorithm Comparisons for Relative Execution Times and Accuracy

slow=24 hours). The data were split into 3 subsets (or time slices) and the preceding 60-80% predicted the full forward time sequence of failures. It is worth noting that the order of events gets rolled into statistical features (mean, standard deviations) and counts (errors, replacements, and machine age).

In the original work, Ref. 5 showed a single regression algorithm (GBM, or gradient boosting machine) and achieved 88-99% recall rates, a measure of sensitivity and specificity that takes account of the unbalanced or dominant normal behavior compared to predicted failures. Recall rates also can highlight the false negative problem, since missed failures can cost more than accepting false positives. To speed up the exploration of algorithms, we under-sampled normal behaviors and balanced the prediction problem from the outset, so that accuracy and time give a simplified performance picture.

## RESULTS

Because the success of any algorithm suite such as Ref.6 will depend on the features or factors devised, we also set out to apply a traditional random forest algorithm and rank order the predictive importance of each input variable to failure rates. As shown in Fig. 2, this importance rank often proves more instructive to the machine learning models than accuracy and time, since the modeler can generalize to find new and better input features. As one might hypothesize initially, the counts for precursor errors gives the most information to the predictive maintenance task, followed by the rotational mean values. The least significant predictors in Fig. 2 included machine age, model and time since last component replacements, all three of which intuitively might otherwise guide a more conventional scheduled maintenance approach.

As shown in Fig. 3, we grouped and sorted classifier results by algorithm method, median accuracy and relative execution time (seconds). We verified the broad success found in Ref. 5 for gradient boosting (shown as gbm), but also identified 3 more accurate (> 96%) and efficient algorithms (Mixture Discriminant Analysis or mda, Penalized Discriminant Analysis or pda, and Multivariate Adaptive Regression Splines or gcvEarth). For the case of penalized discriminants (pda) in Fig. 3, the solution appeared 20 times faster than the previously reported gradient boosting (gbm).

## SUMMARY

This work expands on one of the largest public predictive maintenance dataset that offers a validity test for developing new algorithms, pre-processing protocols for finding relevant features and tuning for the most efficient and accurate outcomes. The strategy of combining many algorithms has proven notably successful in other data competitions, presumably because of its resistance to over-fitting. A corollary to this success hinges on combining enough different approaches to resist adversarial attacks with small alterations in data leading to wildly divergent forecasting results. Future work will expand the classifier suite to include promising but under-represented families such as deep learning and methods for handling unlabeled data such as nearest neighbors and clustering.

## ACKNOWLEDGMENTS

The author would like to thank the PeopleTec Technical Fellows program for encouragement and project assistance. This research benefited from support from U.S. Army Aviation Engineering Directorate, Redstone Arsenal, Alabama and the Special Operation Aviation Regiment (SOAR) Health Usage Maintenance (HUMS) Sustainment program, Ft. Campbell, KY.